# TamilEmo: Finegrained Emotion Detection Dataset for Tamil

Charangan Vasantharajan[1], Sean Benhur[2], Prasanna Kumar Kumarasen[3],
Rahul Ponnusamy[3], Sathiyaraj Thangasamy[4], Ruba Priyadharshini[5],
Thenmozhi Durairaj[6], Kanchana Sivanraju[2], Anbukkarasi Sampath,[7]
Bharathi Raja Chakravarthi[8] John Phillip McCrae[8]
University of Moratuwa, Colombo, Sri Lanka[1]
PSG College of Arts and Science,Coimbatore,India[2]
Indian Institute of Information Technology and Management-Kerala, India.[3]
Sri Krishna Adithya College of Arts and Science, Coimbatore, India.[4]
Sri Sivasubramaniya Nadar College of Engineering, Chennai, India.[6]
ULTRA Arts and Science College, Madurai, India.[5]
Kongu Engineering College, Erode, India.[7]
Insight SFI Research Centre for Data Analytics, National University of Ireland ,Galway.[8]
charangan.18@cse.mrt.ac.lk, seanbenhur@gmail.com, prasanna.mi20@iiitmk.ac.in,rahul.mi20@iiitmk.ac.in,
sathiyarajt@skacas.ac.in,rubapriyadharshini.a@gmail.com,theni_d@ssn.edu.in,kanchana@psgcas.ac.in
anbu.1318@gmail.com, bharathi.raja@insight-centre.org,john.mccrae@insight-centre.org

**Abstract**

Emotional Analysis from textual input has been considered both a challenging and interesting task in Natural Language Processing. However due to the lack of datasets in low-resource languages (i.e. Tamil), it is difficult to conduct research of high standard in this area. Therefore we introduce this labeled dataset (a largest manually annotated dataset of more than 42k Tamil YouTube comments, labeled for 31 emotions including neutral) for emotion recognition. The goal of this dataset is to improve emotion detection in multiple downstream tasks in Tamil.We have also created three different groupings of our emotions (3-class, 7-class and 31-class) and evaluated the model's performance on each category of the grouping. Our MURIL- base model has achieved a 0.60 macro average F1-score across our 3-class group dataset. With 7-class and 31-class groups, the Random Forest model performed well with a macro average F1-scores of 0.42 and 0.29 respectively.

## 1. Introduction

Emotional analysis is the classification task of mining emotions in texts, which finds use in various natural language applications such as reviews analysis in e-commerce, public opinion analysis, extensive search, personalized recommendation, healthcare, and online teaching (Peng et al., 2021).Emotions are psychological state that are often expressed through behaviour or language. Emotional analysis helps to analyze the text and extract the emotions which are expressed in the text. This is why emotional analysis plays a significant role in Natural Language Processing (NLP) and its applications.

Most e-commerce and social media platforms have no restriction on users to creating informal content. Moreover, these platforms allow users to communicate in such a way that they feel comfortable either using native language or switching between one or more languages in the same conversation (Vyas et al., 2014) in order to improve user experience. However, most NLP systems that are trained on monolingual text fail when it comes to analyzing user-generated data (Chakravarthi et al., 2020b). Furthermore, most of the developments in emotional analysis systems are performed on monolingual data for high-resource languages. In contrast, the user-generated content in under-resourced languages is often mixed with English or other high-resource languages (Jose et al., 2020).

Code-Mixed or Code-Switched text is a text that contains mixed vocabulary and syntax of multiple languages at the level of the document, paragraph, comments, sentence, phrase, word or morpheme. (Chakravarthi et al., 2021). Since monolingual text lacks expressions, multilingual societies use codemixed text to their conversation or dialogue (Vasantharajan and Thayasivam, 2021a). Other than the code-mixing, people tend to use roman characters for texting on social media platforms instead of the native script. This has raised many challenges for natural language processing and encouraged researchers to develop various algorithms to handle code-mixed text and roman characters (Vasantharajan and Thayasivam, 2021b).

In this paper, we describe the creation of a monolingual corpus for Tamil (a Dravidian language). Tamil is an official language in Sri Lanka and Singapore. Moreover, Tamil is widely spoken by millions of people around the world. But, the tools and resources available for developing robust NLP applications are under-developed for Tamil. In Tamil, the most complex words are formed by stringing together morphemes without changing them in spelling or phonetics (Sakuntharaj and Mahesan, 2016). Also, the writing system is a phonemic abugida written from left to right, and at one point in history, it used Tamili, Vattezhuthu, Chola, Pallava and Chola-Pallava scripts. Today's modern Tamil script was retrieved from the Chola-Pallava script that was conceived around the 4th century CE (Sakuntharaj and Mahesan, 2018).

Regarding the Dravidian languages, there are various monolingual datasets available for numerous research aims. However, there have been few attempts to make code-mixed datasets for Tamil, Kannada and Malayalam (Chakravarthi et al., 2020a) and (Chakravarthi et al., 2020b). We will help overcome this resource bottleneck for Tamil by creating this dataset in order to equip Tamil with NLP support in emotional analysis in a way that is both cost-effective and rapid. To create resources for a Tamil scenario, we collected comments on various Tamil movie trailers and teasers from YouTube.

The contributions of this paper are:

1. We present a dataset for Tamil for the emotional analysis task.

2. This monolingual dataset contains 31 types of labels with 42,686 comments.

3. We evaluated this dataset on traditional Machine learning models and also provided benchmark results on pre-trained models.

## 2. Related Work

### 2.1. Datasets for Emotion Detection

In the past decade, several researchers have released emotion recognition datasets in several languages for a variety of domains, but they contained only a very small number of emotions, about 4 to 7, also it contained a smaller number of samples. Recently (Rashkin et al., 2019) released an Empathetic dialogues dataset which consisted of text conversations labeled with 32 emotions and (Demszky et al., 2020) released GoEmotions which consists of Reddit comments labeled with 27 emotion categories for around 52k samples. To the best of our knowledge, there are no previous emotion recognition datasets available for Tamil. This is the first work to create a large dataset of fine-grained emotions in the Tamil language.

### 2.2. Emotion Classification

Emotion classification is a type of text classification task that is used to detect emotions. Both traditional machine learning and Deep Learning models can be employed to classify emotions. Traditional machine learning models often use lexicons such as the Valence Arousal Dominance Lexicon (Mohammad, 2018). Deep Learning Models such as RNN, LSTMs and Transformer models can also be employed for this task. Transfer learning methods such as BERT (Devlin et al., 2019), Roberta(Liu et al., 2019) has achieved state of the art results in major NLP tasks. (Cortiz, 2021) fine-tuned different pre-trained models such as BERT, DistilBert, Roberta, XLNet and Electra and Roberta provided the better results. (Suresh and Ong, 2021) and (Khanpour and Caragea, 2018) used a method to incorporate knowledge from emotion lexicons into an attention mechanism to improve emotion classification. In our work, we have used traditional Machine learning models and multilingual transformer models to provide a strong baseline to this dataset.

## 3. Tamil Emotion Dataset

### 3.1. Scraping Raw Data

Data in social media platforms such as Twitter, Facebook and YouTube, are changing rapidly and can alter the perception or reputation of an individual or community drastically. This highlights the importance of automation in emotional analysis. YouTube is a popular social media platform in the Indian subcontinent due to its wide range of content available, such as movies, trailers, songs, tutorials, product reviews and etc. YouTube allows users to create content as well as allows other users to engage and interact with this content through actions such as liking or commenting. Because of this, there is more user-generated content in under-resourced languages.

Hence, we chose YouTube to extract comments to create our dataset. We collected data from movie trailers and teasers to create our dataset because movies are quite popular among the Tamil speaking populace in most countries. This significantly increases the chance of getting various views on one topic from different people. Figure 1 shows the steps that we followed to create our dataset.

We used a YouTube Comment Scraper tool[1] for crawling comments from the YouTub videos and processing the collected comments to make the datasets for emotional analysis with manual annotations. We intended to collect comments that contain purely monolingual texts in the native Tamil language, with enough representation for each emotion class, so we did not worry about Code-Mixed text. As a part of the preprocessing steps to clean the data, we utilized **langdetect** library[2] to capture different languages and to eliminate the unintended languages. Also, in order to maintain data privacy, we made sure that all the user-related information was removed from the corpora. As a part of the text-preprocessing, we removed redundant information such as URL, comments that are nearly the same as other comments according to the **Levenshtein Distance** using a python library[3]. Moreover, we retained all the emojis, emoticons, and punctuation, and etc that are presented in the text to faithfully preserve the real-world usage.

---
[1] https://youtubecommentsdownloader.com/
[2] https://pypi.org/project/langdetect/

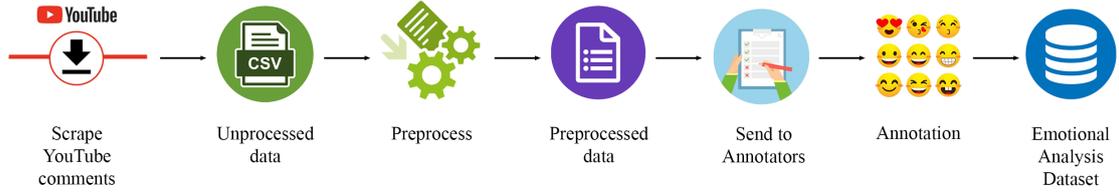

Figure 1: Overview of the data collection process.

### 3.1.1. Methodology of Annotation

We contacted the Indian Institute of Information Technology and Management in order to find volunteers for the annotation process. We created Google Forms that contain 25 comments in one page to collect annotations from annotators and send them through email. The student volunteer annotators received the link to the Google Forms and did the annotations on their personal computers or mobile phones. During the annotation process, gender, educational background and medium of schooling were also collected to ensure that the group of volunteers was diverse, with an equal/adequate reprsentation of gender, sexual orientation and socio-economic backgrounds. The annotators were cautioned that the user suggestions may have aggressive language. They were given a provision to discontinue the annotation process in case the content is too upsetting to deal with.

They were requested not to be partial to a specific individual, circumstance or occasion during the annotation process. The annotators were also instructed to agree on the annotation guidelines before they were allowed to proceed further. We finished the annotation process in three stages. As the first step, each comment was annotated by two individuals. Then included the annotated comment in the dataset if both the annotations agreed. In the event of contention, a third annotator was requested to annotate the comment. In the final stage, if all three of them disagreed, at that point, two additional annotators were asked to label the comments. Therefore, each form was annotated by at least three annotators.

---

[3] https://pypi.org/project/pylev/

### 3.2. Types of Emotions

For emotional analysis, we followed the methodology used by (Chakravarthi et al., 2020b), and included at least three annotators to label each comment. The following annotation guidelines were given to the annotators in Tamil and English.

- **Admiration** (போற்றுதல்): Comments containing respect and warm approval.

- **Amusement** (கேளிக்கை): Comments containing the state or experience of finding something funny.

- **Anger** (கோபம்): Comments containing a strong feeling of being upset or annoyed because of something wrong or bad.

- **Annoyance** (எரிச்சல்): Comments containing the feeling or state of being annoyed; irritation.

- **Anticipation** (எதிர்பார்ப்பு): Comments containing the act of looking forward especially, to expect something to happen / excitement about the future.

- **Approval** (ஒப்புதல்): Comments containing the action of approving something.

- **Caring** (அக்கறை): Comments containing kindness and concern for others.

- **Confusion** (குழப்பம்): Comments that are uncertain about what is happening, intended or required.

- **Curiosity** (ஆர்வம்): Comments containing a strong desire to know or learn something.

- **Desire** (ஆசை): Comments containing a strong feeling of wanting to have something or wishing for something to happen.

- **Disappointment** (ஏமாற்றம்): Comments containing sadness or displeasure caused by the non-fulfilment of one's hopes or expectations.

- **Disapproval** (மறுப்பு): Comments containing possession or expression of an unfavourable opinion.

- **Disgust** (அருவருப்பு): Comments containing a strong feeling of dislike for something that has a very unpleasant appearance, taste, smell, etc.

- **Embarrassment** (சங்கடம்): Comments containing a feeling of self-consciousness, shame, or awkwardness.

- **Excitement** (உற்சாகம்): Comments containing a feeling of great enthusiasm and eagerness.

- **Fear** (பயம்): Comments containing a feeling of anxiety and agitation caused by the presence or nearness of danger, evil, pain, etc.

- **Gratitude** (நன்றியறிதல்): Comments containing the quality of being thankful; readiness to show appreciation for and to return kindness.

- **Grief** (துக்கம்): Comments containing intense sorrow, especially caused by someone's death.

- **Joy** (மகிழ்ச்சி): Comments containing a feeling of great pleasure and happiness.

- **Love** (அன்பு): Comments containing a feeling of strong or constant affection for a person.

- **Nervousness** (பதட்டம்): Comments containing the quality or state of being nervous.

- **Neutral** (நடுவுநிலை): Comments that are not supportive or helpful to either side in a conflict, disagreement, etc.

- **Optimism** (எதிர்காலத்தைப் பற்றிய நம்பிக்கை): Comments containing hopefulness and confidence about the future or the success of something.

- **Pride** (பெருமை): Comments containing a feeling of deep pleasure or satisfaction derived from one's own achievements, the achievements of those with whom one is closely associated, or from qualities or possessions that are widely admired.

- **Realization** (உண்மையை உணர்தல்): Comments containing an act of becoming fully aware of something as a fact.

- **Relief** (துயர்நீக்கம்): Comments containing a feeling of reassurance and relaxation following release from anxiety or distress.

- **Remorse** (குற்றமுணர்ந்ததால் ஏற்படும் வருத்தம்): Comments containing deep regret or guilt for a wrong committed.

- **Sadness** (சோகம்): Comments associated with, or characterized by, feelings of disadvantage, loss, despair, grief, helplessness, disappointment and sorrow.

- **Surprise** (ஆச்சரியம்): Comments containing a completely unexpected occurrence, appearance, or statement.

- **Teasing** (கிண்டல்): Comments intended to provoke or make fun of someone in a playful way.

- **Trust** (நம்பிக்கை): Comments containing assured reliance on the character, ability, strength, or truth of someone or something.

### 3.3. Annotator Statistics

Since we called the volunteers from various Universities and Colleges, the types of annotators in the task are different. From Table 1, we can see that 9 male and 7 female annotators volunteered to contribute to the task. The distribution of male annotators is slightly higher than female annotators in the task. The majority of the annotators have received undergraduate levels of education. We were also unable to find volunteers of the transgender community to annotate our dataset. From Table 1, we can observe that the majority of the annotators' medium of schooling is English even though their mother tongue is Tamil. Since most of the participants' medium of education was the English language, we were careful that it would not affect the annotation process by ensuring that all of them are fully proficient in using the Tamil language.

Once the forms were created, a sample form (first assignment) was annotated by experts and a gold standard was created. Then, we manually compared the volunteer's submission form with gold standard annotations. Then, we eliminated the annotators for the following reasons during the first assignment to control the quality of annotation. They are:

| **Language** | | **Tamil** |
|---|---|---|
| Gender | Male | 09 |
| | Female | 07 |
| | Transgender | 00 |
| Higher Education | Undergraduate | 12 |
| | Postgraduate | 04 |
| Higher Education | Undergraduate | 12 |
| | Postgraduate | 04 |
| Medium of Schooling | English | 09 |
| | Native language | 07 |

Table 1: Annotators Statistics for emotional analysis task.

- Annotators delayed unreasonably for responding.
- labeled all sentences with the same label.
- More than 10 annotations in a form were wrong.

After the elimination, a total of 16 volunteers were involved in the annotation task. Once they completed the given Google Forms, 10 Google Forms (250 comments) were sent to them. If an annotator offered to volunteer more, the next set of Google Forms was sent to them with another set of 250 comments and in this way, each volunteer chose to annotate as many comments from the corpus as they wished.

### 3.4. Inter-annotator agreement

The inter-annotator agreement is the measure of how well two or more annotators can make the same annotation decision for a certain category. This is important to verify that the annotation process is consistent and that different annotators are able to label the same emotion label to a given comment. Generally, there are two questions that arise about the inter-annotator agreement. They are:

- How do the raters agree or disagree in their annotation?
- How much of the observed agreement or disagreement among the annotators might be due to chance?

To answer these questions and to ensure the annotation reliability, we used Krippendorff's alpha($\alpha$)(Krippendorff, 1970) as a inter-annotator agreement measure. For instance, if the annotators vary between Admiration and Anticipation classes, the difference is more genuine than when they differ between Optimism and Trust classes. $\alpha$ is sensitive and is characterized as follows:

$$\alpha = 1 - \frac{D_o}{D_e}$$

where $D_o$ is the observed disagreement between emotion labels annotated by the annotators and $D_e$ is the expected disagreement when the coding of emotions can be attributed to chance rather than due to the inherent property of the emotion itself. We used **agreement** module from nltk[4] for calculating $\alpha$. The results of inter-annotator agreement between our annotators is 0.7452.

---

[4] https://www.nltk.org/

### 3.5. Selecting & Curating YouTube comments

We choose YouTube videos with at least 500 comments and remove non-English comments. Generally, YouTube comments contain biased data about a particular video. So to reduce biased training data, we process the dataset through a series of data curating methods as below. Therefore, we can develop fully representative emotion models using our dataset and can ensure our data does not comment that are emotion-specific or language biases.

**Manual Review**
We read the comments manually to detect and remove harmful comments towards a particular religion, ethnicity, gender, sexual orientation, or disability, to create the best dataset of our judgment.

**Length Filtering**
We choose comments with 3-512 tokens including punctuation using NLTK's word tokenizer library[5].

**Levenshtein Distance**
Levenshtein distance is a measure to calculate the distance between two strings as the minimum number of characters needed to insert, delete or replace in a given string to transform it to another string. We removed the comments which are having more than 80% similarly by using **pylev**[6].

## 4. Data Analysis

Table 2 contains the summary statistics of this dataset. Figure 2 shows the distribution of emotion labels, we can notice that emotion Admiration holds up the highest number of examples and desire is the lowest count.

| Description | Count |
| --- | --- |
| No of examples | 42686 |
| No of emotions | 31 |
| Max no of tokens in an text | 2988 |
| Min. no of tokens in an text | 2 |
| Max. no of data in a class | 6682-admiration |
| Min. no of data in a class | 208-desire |

Table 2: Dataset Statistics.

### 4.1. Topic Modelling

Topic modelling is an approach for automatically identifying topics in a collection of documents. The most commonly used topic modelling method is Latent Dirichlet Allocation(LDA) (Blei et al., 2003), It represents topics in a document and the distribution of words inside each topic. LDA has been previously used for many use-cases. Related to emotion analysis. (Dakshina and Sridhar, 2014) has used LDA to recognise emotions in the lyrics of songs, (Güven et al., 2019) have used LDA to identify emotions in Turkish tweets.

---

[5] https://www.nltk.org/api/nltk.tokenize.html
[6] https://pypi.org/project/pylev/

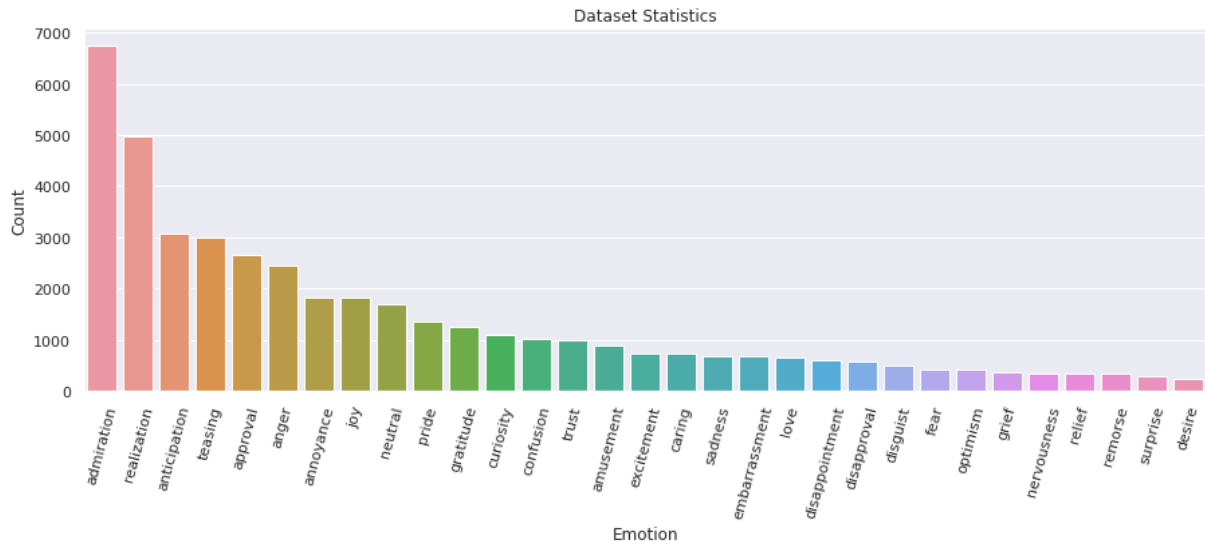

Figure 2: Dataset Statistics

| Topic Keywords | No.of Docs | % of Docs |
|---|---|---|
| உண்மை, ஆனால், அல்ல, தெரியும், வரலாறு, முடியும், சொல்வது, நம்பிக்கை, இதுதான், ஆயிரம் | 3232 | 0.0757 |
| சூப்பர், கதை, படம், படத்தை, தலைவா, சொல்லும், அண்ணன், செம, உண்மையை, ப்ரோ | 2604 | 0.061 |
| ஆட்சி, திமுக, சாமி, தலைவர், ஸ்டாலின், முதல்வர், செய்த, நேதாஜி, தமிழக, ஜெயலலிதா | 2385 | 0.059 |
| வாழ்த்துக்கள், படம், சார்பாக, வெற்றி, பெற, வாழ்த்துகள், ரசிகர்கள், திரைப்படம், அடைய, வெற்றியடைய | 2217 | 0.0519 |
| உங்கள், நீங்கள், உங்களுக்கு, அம்மா, அக்கா, உங்களை, எங்கள், பேச்சு, அழகு, நாங்கள் | 2209 | 0.0517 |
| இல்லை, தமிழர், கடவுள், கட்சி, பெரியார், உண்மையான, வார்த்தை, சரியாக, விடுதலை, சொன்னது | 1918 | 0.0449 |
| பெரிய, தமிழ்நாடு, சரியான, இருக்கிறது, உண்டு, உலகம், சென்று, எவ்வளவு, எப்படி, சோழன் | 1824 | 0.0427 |

Table 3: Dominant keywords associated with each Topics extracted using LDA

We use python's MALLET (McCallum, 2002) and Gensim (Rehurek and Sojka, 2011) package to run the LDA model. we remove punctuations, URLs, emojis and Tamil stopwords[7] since most repetitive words can cause noise in the dataset. In general topic model requires a collection of documents as an input, we perform topic modelling on a document level, with k=8 to cluster the topics. For our analysis, we then extract the most representative topics and provide them in Table 3. The words such as தமிழ்நாடு, சோழன் are grouped together which refers to topics on TamilNadu and tamil historical topics and கட்சி, பெரியார், உண்மையான, வார்த்தை, சரியாக, விடுதலை refers to topics on political parties.

### 4.2. Keywords

We extract the top words associated with emotion category by calculating the log odds ratio, informative Dirichlet prior(Monroe et al., 2017) for each token. The log odds ratio quantifies the association between a word in one emotion category to other emotion categories by building on word frequency counts and an estimate of prior word probability, this helps us to extract the important keywords of each emotion. The log odds are z-scored, all values greater than 1 indicates association with the corresponding emotion. We list the top 5 tokens of each emotion in Table 4. We can note that the keywords are highly correlated with the emotion (e.g amusement with tokens such as ஹேத்துனா,மாங்கல்யம்,, pride with "நிமிர்ந்த"

## 5. Modeling

### 5.1. Data Preparation

After the annotation process, each comment was labeled by at last three annotators. So we filter out emotion labels selected by only a single annotator to reduce the noise in our data. After the cleaning process 97% of the data was retained as the original. Finally, we randomly split the dataset into

---
[7] https://github.com/AshokR/TamilNLP/wiki/Stopwords

| admiration | amusement | anger | annoyance | anticipation | approval | caring |
|---|---|---|---|---|---|---|
| வாழ்த்துக்கள்(30.37) | ஹேத்துனா(3.63) | நாயே(5.04) | ஏண்டா(2.58) | வேண்டும்(8.96) | அல்ல(2.86) | கவலை(1.6) |
| சார்பாக(22.16) | தந்தனானே(3.59) | டேய்(4.54) | வன்னிய(2.53) | போடுங்க(6.11) | இல்லை(2.58) | பனை(1.38) |
| வெற்றி(21.05) | மாங்கல்யம்(3.53) | நாய்களை(4.01) | போங்கடா(2.5) | ஆதரவு(5.4) | உண்மை(2.44) | முடிந்த(1.29) |
| பெற(19.23) | ஜீவன(3.09) | நாய்கள்(4) | ஏன்டா(2.45) | தாருங்கள(4.35) | தான்(2.34) | கவனமாக(1.23) |
| வாழ்க(16.08) | போட்றா(2.78) | திருட்டு(3.76) | அவன்(2.35) | வீடியோ(4.1) | என்பது(2.11) | வேண்டாம்(1.2) |

| confusion | curiosity | desire | disappointment | disapproval | disgust | embarrassment |
|---|---|---|---|---|---|---|
| நல்லவனா(2.22) | ஆர்வமாக(2.44) | ஆவலாய்(1.59) | விடியல்(2.41) | மெய்த்து(2.23) | அருகரீகம்(1.73) | சாயம்(1.3) |
| சந்தேகம்(1.9) | சொல்லுங்க(2.37) | பான்பராக்(1.36) | ரூபாய்(1.94) | ஆரியர்கானான(2.05) | உலுறவு(1.56) | வேதனை(1.1) |
| ஆகலாம்(1.9) | பாகம்(2.22) | பகுத்தரிவாளர்(1.19) | கிலோ(1.86) | கூறுகிறான்(1.89) | பித்து(1.56) | |
| தெரி(1.84) | எப்ப(2.17) | ஆசை(1.18) | பள்ளர்கள்(1.75) | பிரமாணர்கள்(1.81) | பயலே(1.43) | |
| கல்வெட்டு(1.77) | அடுத்த(2.13) | தடவைக்கு(1.07) | தேர்தலுக்கு(1.54) | இல்லை(1.65) | போடா(1.4) | |

| excitement | fear | gratitude | grief | joy | love | nervousness |
|---|---|---|---|---|---|---|
| சூப்பர்(2.65) | பயமா(3.47) | நன்றி(21.42) | ஓடினேன்(1.94) | வாழ்த்துக்கள்(7.55) | அக்கா(3.7) | போய்விட்டார்(2.34) |
| அட்ரா(2.58) | தமிழ்நாடு(2.85) | மிக்க(8.42) | நம்முடிரிகளும்(1.94) | சார்பாக(5.96) | பிடிக்கும்(2.74) | அவர்கள்(1.74) |
| மாஸ்(2.09) | தொண்டை(2.11) | நன்றிகள்(7.01) | நாயர்களும்(1.94) | மகிழ்ச்சி(5.93) | அம்மா(2.53) | ஆலயம்(1.12) |
| கூர்ஸ்ட்(1.89) | இருமல்(1.96) | சகோ(3.79) | பிடிபுகளுக்கு(1.9) | வெற்றி(5.61) | செல்லக்குட்டி(2.19) | |
| அடிடா(1.76) | வாத்திகள்(1.94) | தகவல்களுக்கு(3.3) | கையாலாகாத(1.57) | படம்(4.81) | என்(2) | |

| neutral | optimism | pride | realization | relief | remorse | sadness |
|---|---|---|---|---|---|---|
| மலையாளி(3.58) | சம்பந்தியாக(1.94) | தமிழன்(6.1) | உண்மை(3.29) | னுங்க(1.67) | அவளது(2.06) | ஆயிரத்தில்(3.81) |
| ஆலங்காலங்குருவி(3.12) | உள்நாட்டு(1.44) | பெருமை(5.94) | என்று(3.12) | ஒய்க்கை(1.59) | வாக்கை(1.96) | ஊஹர்(2.89) |
| ஆகாசத்து(3.12) | போரை(1.41) | தமிழ்(5.65) | தான்(3.09) | தமிழ்(1.59) | சரியானதாக(1.94) | ஒருவன்(2.26) |
| செயலாளர்(2.85) | வெல்வோம்(1.36) | நிமிர்ந்து(4.81) | அவர்கள்(2.99) | அனுபவோம்(1.45) | வேலைவாய்ப்பு(1.44) | நட்டேன்(2.23) |
| தழுவி(2.68) | நாம்(1.34) | நில்லடா(4.6) | என்பது(2.85) | குரானை(1.35) | தேவைக்கு(1.44) | கண்ணீர்(1.59) |

| surprise | teasing | trust |
|---|---|---|
| தயாரிச்சது(1.59) | திருப்பி(3.86) | கிருபா(3.17) |
| வாங்கினா(1.59) | மாமா(3.38) | கடவுள்(3.06) |
| ஆச்சரியமாக(1.45) | குருமா(3.33) | நம்புகிறேன்(2.01) |
| ஆச்சரியம்(1.38) | கொடுக்காத(3.25) | நல்லதே(1.99) |
| செய்தியா(1.19) | விருதை(3.14) | இறைவனை(1.73) |

Table 4: Top 5 words associated with each emotion. The log odds ratios are showed in the parentheses.

train(80%), Dev(10%), Test(10%) sets. Once our model was finalized, Dev and Test sets were used to validate and evaluate our model respectively.

**Grouping emotions**
We create three different groups of our taxonomy and evaluated the model's performance on each category of the grouping. A emotion level(Group 01) divides the labels into 3 categories - positive(நேர்மறை), negative(எதிர்மறை) and ambiguous(தெளிவற்ற).A further grouping divides the labels into 7 categories. They are love-அன்பு(maps to: caring, love, anticipation, excitement, joy, desire, and curiosity), Pathos-அழுகை(maps to: disgust, embarrassment, sadness, grief, relief, and remorse), Disgrace-இளிவரல்(maps to: annoyance, disappointment, anger, nervousness, fear, confusion, and disapproval), Laughter-நகை(maps to: teasing, and amusement), neutral-நடுவுநிலை(maps to: realization, approval, and neutral), hope-நம்பிக்கை(maps to: trust, pride, admiration, and optimism), and Bewilderment-மருட்கை(maps to: surprise and gratitude). Figure 3 & 4 show the mapping between the emotion classes.

### 5.2. Baseline Experiments

In this section, we will describe our methodology for building a baseline system for developing an Emotion detection system for the created dataset. We experiment with both traditional Machine Learning models and pretrained multilingual models such as mBERT and MURIL.

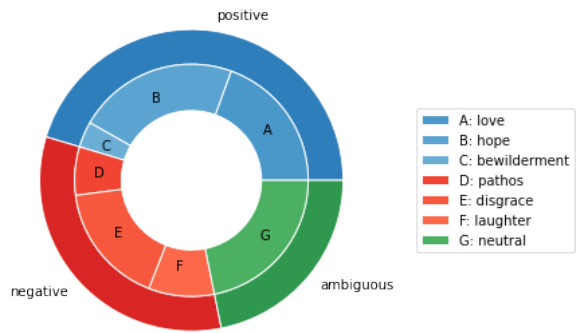

Figure 3: 3-Class Group Mapping

#### 5.2.1. Machine Learning Models
We train traditional Machine learning models as a baseline for this dataset. We experimented with Logistic Regression, K-Nearest Neighbors classifier, Linear Support Vector Machines(SVM) (Cortes and Vapnik, 1995), Decision Trees, Random Forest. For all the models we encode the text into TF-IDF vectors of uni-grams and pass it to the respective models.

#### 5.2.2. mBERT
This model was pre-trained using the same pre-training strategy that was employed to BERT(Devlin et al., 2019), which is Masked Language

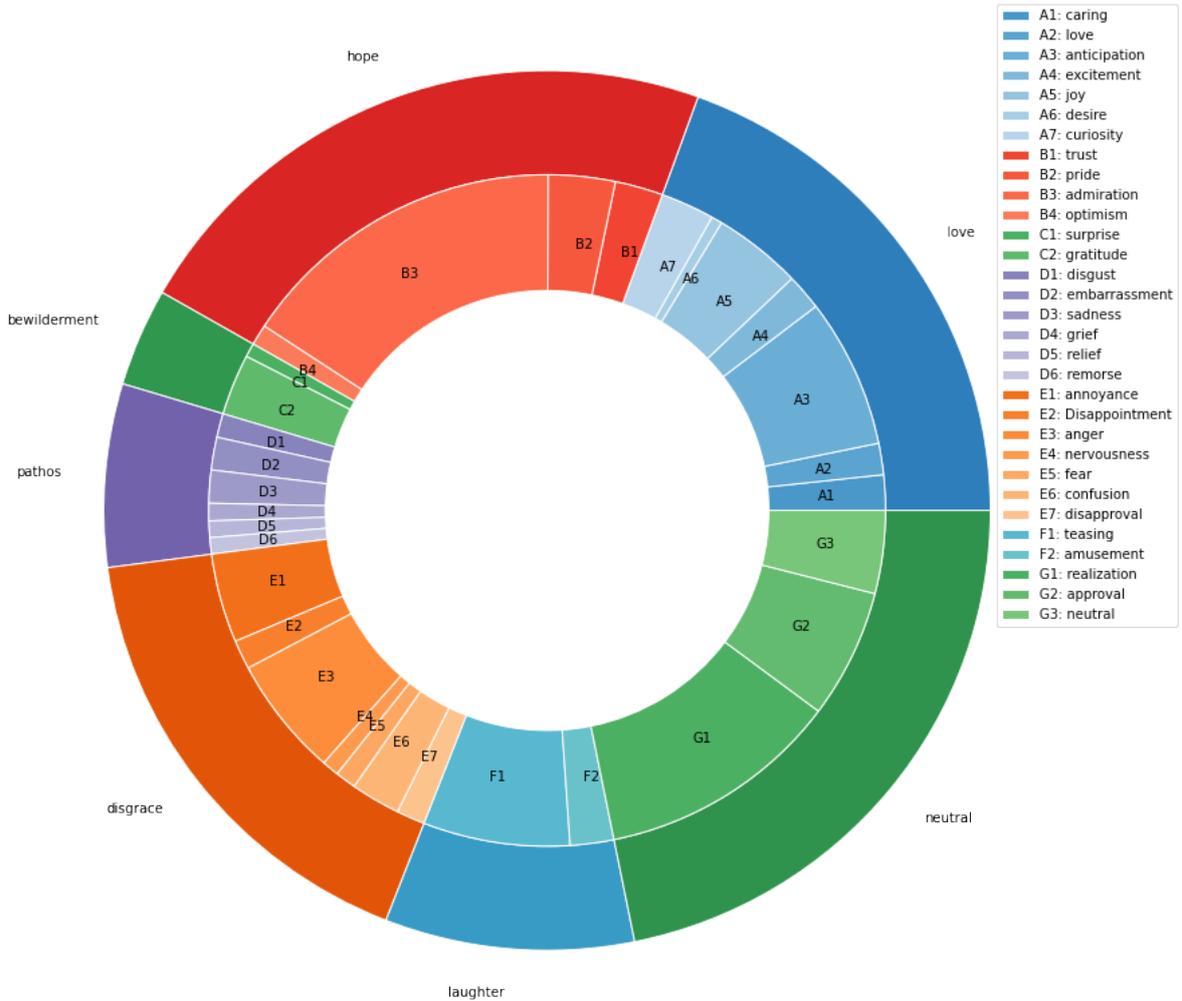

Figure 4: 7-Class Group Mapping

| Model | Parameters |
| --- | --- |
| Logistic Regression | optimizer='lbfgs', penalty='l2' |
| Support Vector Machine | max_iter=2000, kernel='rbf', tol=0.0001 |
| Multinomial Naive Bayes | alpha=1.0, class_prior=None, fit_prior=None |
| Random Forest | criterion='gini', n_estimators=100 |
| K Nearest Neighbors | n_neighbors=5, algorithm='brute' |

Table 5: Hyperparameters for Machine Learning Models

Modelling (MLM) and Next Sentence Prediction (Next Sentence Prediction). It was pre-trained on the Wikipedia dump of 104 languages. For having a balanced set of all languages exponentially smoothed weighting of data was performed during data creation and vocabulary creation. This results in high resource languages being under-sampled while low resourced languages are being over-sampled.

#### 5.2.3. MuRIL
MuRIL(Khanuja et al., 2021) is a pre-trained model, trained on 16 different Indian Languages and English languages, it was trained on both the MLM and Translated Language Modelling(TLM) objectives. The model was trained with Common crawl OSCAR corpus and Wikipedia corpus.

### 5.3. Experiment Settings
We removed the unwanted noises in the text such as URLs, punctuation, digits, unwanted symbols and emojis during the preprocessing step.

| Model | 3-class Group | | | 7-classs Group | | | 31-class Group | | |
|---|---|---|---|---|---|---|---|---|---|
| | P | R | F1 | P | R | F1 | P | R | F1 |
| Logistic Regression | 0.54 | 0.53 | 0.52 | 0.35 | 0.36 | 0.32 | 0.17 | 0.18 | 0.16 |
| Support Vector Machine | 0.54 | 0.53 | 0.52 | 0.39 | 0.35 | 0.33 | 0.24 | 0.14 | 0.14 |
| K-Nearest Neighbors | 0.51 | 0.50 | 0.50 | 0.37 | 0.35 | 0.35 | 0.30 | 0.21 | 0.23 |
| Multinomial Naive Bayes | 0.54 | 0.49 | 0.48 | 0.49 | 0.28 | 0.28 | 0.11 | 0.07 | 0.05 |
| Decision Tree | 0.42 | 0.41 | 0.41 | 0.23 | 0.24 | 0.23 | 0.11 | 0.08 | 0.08 |
| Random Forest | 0.61 | 0.55 | 0.56 | 0.49 | 0.41 | 0.42 | 0.47 | 0.25 | 0.29 |
| mBERT | 0.56 | 0.56 | 0.55 | 0.36 | 0.40 | 0.36 | 0.12 | 0.15 | 0.10 |
| MuRIL-Base | 0.60 | 0.60 | 0.60 | 0.25 | 0.32 | 0.25 | 0.04 | 0.07 | 0.03 |

Table 6: Macro Average Precision, Recall and F1-Score on test set for all the Models

| Sentiment | Precision | Recall | F1 |
|---|---|---|---|
| positive | 0.76 | 0.70 | 0.73 |
| negative | 0.58 | 0.51 | 0.54 |
| ambiguous | 0.39 | 0.53 | 0.45 |
| **macro avg** | **0.60** | **0.60** | **0.60** |

Table 7: Results of MuRIL on sentiment-grouped taxonomy

| Emotion | Precision | Recall | F1 |
|---|---|---|---|
| Hope | 0.47 | 0.67 | 0.55 |
| Neutral | 0.48 | 0.44 | 0.46 |
| Love | 0.37 | 0.44 | 0.40 |
| Bewilderment | 0.61 | 0.41 | 0.49 |
| Disgrace | 0.47 | 0.39 | 0.43 |
| Pathos | 0.65 | 0.20 | 0.30 |
| Laughter | 0.40 | 0.29 | 0.34 |
| **macro avg** | **0.49** | **0.41** | **0.42** |

Table 8: Results of Random Forest on 7-class taxonomy

**Machine Learning Models**

We trained all the models with uni-grams extracted from the TF-IDF technique. All the parameters are summarized in Table 5. In Logistic Regression the lbfgs optimizer was used with l2 regularizer, a maximum of 2000 iterations is taken for a solver to converge. For SVM we have used rbf kernel for a maximum of 2000 iterations with a tolerance of 1e-3. In the Random Forest model, we used 100 trees with gini criterion, which is used to measure the quality of the split in the tree. The additive feature of Naive Bayes is set to 1 and classprior and fitprior is set to None. For the Decision Tree, a gini criterion was used to split the tree with the best split based on node. The hyperparameters used during the training is shown in Table 5.

**Pre-trained Models**

We fine-tuned mBERT and MuRIL pre-trained models for 5 epochs with a batch size of 8, linear scheduler with 2e-5 as an initial learning rate is used, the models are trained with AdamW optimizer and a Cross-Entropy loss function. We found that training for more epochs causes overfitting of the model.

### 5.4. Results and Discussion

Table 6 summarizes the results of all the models on the different groupings of the data. Random Forest achieves the best macro average F1-score of 0.29 for the full taxonomy and macro average F1 of 0.42 for the seven class taxonomy.

The MuRIL-Base model achieves a macro average of 0.60 for the sentiment grouped taxonomy but it achieves a very less score for the full taxonomy. Table 7, 8 and 9 summarizes the class-wise results of the best performing models Random Forest and MuRIL on sentiment taxonomy, seven class taxonomy, and full taxonomy respectively. The model achieves the highest F1-score for Admiration which contains the most number of samples in the dataset and it achieves the lowest score of 0.16 for grief.

After grouping the data, MuRIL achieves the highest macro average F1 Score of 0.60 in sentiment level taxonomy and Random Forest achieves 0.36 macro average F1-Score on seven-class taxonomy. This shows that it is significantly hard for ML models to compare emotions in fine-grained settings.

### 6. Conclusion

We propose TamilEmo, a large dataset for fine-grained emotion detection that has been extensively annotated manually. We present a detailed data analysis that demonstrates the accuracy of the annotations over the whole taxonomy with a high inter-annotator agreement in terms of Krippendorff's alpha. Also, for each class group, we give the results of all models in terms of precision, recall, and F-Score. We anticipate that this resource will enable researchers to tackle new

| Emotion | Precision | Recall | F1 |
|---|---|---|---|
| admiration | 0.32 | 0.76 | 0.45 |
| amusement | 0.61 | 0.18 | 0.18 |
| anger | 0.41 | 0.28 | 0.34 |
| annoyance | 0.39 | 0.23 | 0.29 |
| anticipation | 0.36 | 0.29 | 0.32 |
| approval | 0.41 | 0.26 | 0.31 |
| caring | 0.57 | 0.21 | 0.31 |
| confusion | 0.39 | 0.19 | 0.25 |
| curiosity | 0.35 | 0.18 | 0.24 |
| desire | 0.45 | 0.22 | 0.29 |
| disappointment | 0.73 | 0.23 | 0.35 |
| disapproval | 0.41 | 0.14 | 0.21 |
| disgust | 0.50 | 0.18 | 0.26 |
| embarrassment | 0.73 | 0.15 | 0.25 |
| excitement | 0.50 | 0.15 | 0.23 |
| fear | 0.82 | 0.33 | 0.47 |
| gratitude | 0.55 | 0.56 | 0.55 |
| grief | 0.50 | 0.10 | 0.16 |
| joy | 0.38 | 0.23 | 0.29 |
| love | 0.18 | 0.18 | 0.18 |
| nervousness | 0.50 | 0.12 | 0.20 |
| neutral | 0.39 | 0.20 | 0.26 |
| optimism | 0.44 | 0.16 | 0.23 |
| pride | 0.30 | 0.32 | 0.31 |
| realization | 0.38 | 0.42 | 0.40 |
| relief | 0.60 | 0.19 | 0.29 |
| remorse | 0.33 | 0.12 | 0.18 |
| sadness | 0.64 | 0.24 | 0.35 |
| surprise | 0.43 | 0.22 | 0.29 |
| teasing | 0.25 | 0.43 | 0.32 |
| trust | 0.61 | 0.18 | 0.27 |
| **macro avg** | **0.47** | **0.25** | **0.29** |

Table 9: Results of Random Forest on Full Taxonomy

and fascinating Tamil research within the topics of emotion detection.

## 7. Bibliographical References

Blei, D. M., Ng, A. Y., and Jordan, M. I. (2003). Latent dirichlet allocation. *J. Mach. Learn. Res.*, 3(null):993–1022, mar.

Chakravarthi, B. R., Jose, N., Suryawanshi, S., Sherly, E., and McCrae, J. P. (2020a). A sentiment analysis dataset for code-mixed malayalam-english. In *SLTU*.

Chakravarthi, B. R., Muralidaran, V., Priyadharshini, R., and McCrae, J. P. (2020b). Corpus creation for sentiment analysis in code-mixed Tamil-English text. In *Proceedings of the 1st Joint Workshop on Spoken Language Technologies for Under-resourced languages (SLTU) and Collaboration and Computing for Under-Resourced Languages (CCURL)*, pages 202–210, Marseille, France, May. European Language Resources association.

Chakravarthi, B. R., Priyadharshini, R., Muralidaran, V., Jose, N., Suryawanshi, S., Sherly, E., and McCrae, J. P. (2021). Dravidiancodemix: Sentiment analysis and offensive language identification dataset for dravidian languages in code-mixed text.

Cortes, C. and Vapnik, V. (1995). Support-vector networks. *Machine learning*, 20(3):273–297.

Cortiz, D. (2021). Exploring transformers in emotion recognition: a comparison of bert, distillbert, roberta, xlnet and electra.

Dakshina, K. and Sridhar, R. (2014). Lda based emotion recognition from lyrics. In Malay Kumar Kundu, et al., editors, *Advanced Computing, Networking and Informatics- Volume 1*, pages 187–194, Cham. Springer International Publishing.

Demszky, D., Movshovitz-Attias, D., Ko, J., Cowen, A., Nemade, G., and Ravi, S. (2020). GoEmotions: A dataset of fine-grained emotions. In *Proceedings of the 58th Annual Meeting of the Association for Computational Linguistics*, pages 4040–4054, Online, July. Association for Computational Linguistics.

Devlin, J., Chang, M.-W., Lee, K., and Toutanova, K. (2019). Bert: Pre-training of deep bidirectional transformers for language understanding.

Güven, Z., Diri, B., and Cakaloglu, T. (2019). Emotion detection with n-stage latent dirichlet allocation for turkish tweets. *Academic Platform Journal of Engineering and Science*, pages 467–472, 09.

Jose, N., Chakravarthi, B. R., Suryawanshi, S., Sherly, E., and McCrae, J. P. (2020). A survey of current datasets for code-switching research. *2020 6th International Conference on Advanced Computing and Communication Systems (ICACCS)*, pages 136–141.

Khanpour, H. and Caragea, C. (2018). Fine-grained emotion detection in health-related online posts. In *Proceedings of the 2018 Conference on Empirical Methods in Natural Language Processing*, pages 1160–1166, Brussels, Belgium, October-November. Association for Computational Linguistics.

Khanuja, S., Bansal, D., Mehtani, S., Khosla, S., Dey, A., Gopalan, B., Margam, D. K., Aggarwal, P., Nagipogu, R. T., Dave, S., Gupta, S., Gali, S. C. B., Subramanian, V., and Talukdar, P. (2021). Muril: Multilingual representations for indian languages.

Krippendorff, K. (1970). Estimating the reliability, systematic error and random error of interval data. *Educational and Psychological Measurement*, 30(1):61–70.

Liu, Y., Ott, M., Goyal, N., Du, J., Joshi, M.,

Chen, D., Levy, O., Lewis, M., Zettlemoyer, L., and Stoyanov, V. (2019). Roberta: A robustly optimized bert pretraining approach.

McCallum, A. K. (2002). Mallet: A machine learning for language toolkit. http://www.cs.umass.edu/ mccallum/mallet.

Mohammad, S. M. (2018). Obtaining reliable human ratings of valence, arousal, and dominance for 20,000 english words. In *Proceedings of The Annual Conference of the Association for Computational Linguistics (ACL)*, Melbourne, Australia.

Monroe, B. L., Colaresi, M. P., and Quinn, K. M. (2017). Fightin' words: Lexical feature selection and evaluation for identifying the content of political conflict. *Political Analysis*, 16(4):372–403.

Peng, S., Cao, L., Zhou, Y., Ouyang, Z., Yang, A., Li, X., Jia, W., and Yu, S. (2021). A survey on deep learning for textual emotion analysis in social networks. *Digital Communications and Networks*.

Rashkin, H., Smith, E. M., Li, M., and Boureau, Y.-L. (2019). Towards empathetic open-domain conversation models: a new benchmark and dataset.

Rehurek, R. and Sojka, P. (2011). Gensim–python framework for vector space modelling. *NLP Centre, Faculty of Informatics, Masaryk University, Brno, Czech Republic*, 3(2).

Sakuntharaj, R. and Mahesan, S. (2016). A novel hybrid approach to detect and correct spelling in tamil text. *2016 IEEE International Conference on Information and Automation for Sustainability (ICIAfS)*, pages 1–6.

Sakuntharaj, R. and Mahesan, S. (2018). Detecting and correcting real-word errors in tamil sentences. *Ruhuna Journal of Science*.

Suresh, V. and Ong, D. C. (2021). Using knowledge-embedded attention to augment pre-trained language models for fine-grained emotion recognition.

Vasantharajan, C. and Thayasivam, U. (2021a). Hypers@DravidianLangTech-EACL2021: Offensive language identification in Dravidian code-mixed YouTube comments and posts. In *Proceedings of the First Workshop on Speech and Language Technologies for Dravidian Languages*, pages 195–202, Kyiv, April. Association for Computational Linguistics.

Vasantharajan, C. and Thayasivam, U. (2021b). Towards offensive language identification for tamil code-mixed youtube comments and posts. *SN Computer Science*, 3(1):94.

Vyas, Y., Gella, S., Sharma, J., Bali, K., and Choudhury, M. (2014). Pos tagging of english-hindi code-mixed social media content. In *EMNLP*.